\documentclass[10pt,twocolumn,letterpaper]{article}

\usepackage{cvpr}              

\usepackage[accsupp]{axessibility}  
\usepackage{multirow}
\usepackage{url}
\usepackage{wrapfig}
\usepackage{bm}
\usepackage{colortbl}
\usepackage{listings}
\usepackage{xcolor}

\definecolor{cvprblue}{rgb}{0.21,0.49,0.74}
\usepackage[pagebackref,breaklinks,colorlinks,allcolors=cvprblue]{hyperref}

\def\paperID{***} 
\def\confName{CVPR}
\def\confYear{2025}

\title{ShiftwiseConv: Small Convolutional Kernel with Large Kernel Effect}

\author{
	{Dachong Li}$^{1}$
	~~~ {Li Li}$^{2}$
	~~~ {Zhuangzhuang Chen}$^{1}$
	~~~ {Jianqiang Li}$^{3}$\thanks{Corresponding author. This work is supported in part by the National Natural Science Funds for Distinguished Young Scholar under Grant 62325307, in part by the National Natural Science Foundation of China under Grants 62203134, in part by the Natural Science Foundation of Guangdong Province under Grants 2023B1515120038, in part by Shenzhen Science and Technology Innovation Commission (20220809141216003, KJZD20230923113801004), in part by the Scientific Instrument Developing Project of Shenzhen University under Grant 2023YQ019. } \\
	\textsuperscript{1} College of Computer Science and Software Engineering, Shenzhen University, Shenzhen, China  \\ 
        \textsuperscript{2} School of Mathematical Sciences, Shenzhen University, Shenzhen, China  \\ 
	~~\textsuperscript{3}~ National Engineering Laboratory for Big Data System Computing Technology. \\ 
        \texttt{\small{\{lidachong2023, 
                        2250201001,
                        chenzhuangzhuang2016\}@email.szu.edu.cn,
                        lijq@szu.edu.cn
}}}

\begin{document}

\maketitle
\begin{abstract}

Large kernels make standard convolutional neural networks (CNNs) great again over transformer architectures in various vision tasks. 
Nonetheless, recent studies meticulously designed around increasing kernel size have shown diminishing returns or stagnation in performance.
Thus, the hidden factors of large kernel convolution that affect model performance remain unexplored. In this paper, we reveal that the key hidden factors of large kernels can be summarized as two separate components: extracting features at a certain granularity and fusing features by multiple pathways. To this end, we leverage the multi-path long-distance sparse dependency relationship to enhance feature utilization via the proposed Shiftwise (SW) convolution operator with a pure CNN architecture. In a wide range of vision tasks such as classification, segmentation, and detection, SW surpasses state-of-the-art transformers and CNN architectures, including SLaK and UniRepLKNet. More importantly, our experiments demonstrate that $3 \times 3$ convolutions can replace large convolutions in existing large kernel CNNs to achieve comparable effects, which may inspire follow-up works. Code and all the models at  \url{https://github.com/lidc54/shift-wiseConv}.

\end{abstract}    
\section{Introduction}
\label{sec:intro}

Vision Transformer (ViT)~\cite{liu2021swin, dong2022cswin,li2024multi,li2024spiking} outperforms traditionally dominant CNNs, largely due to its superior capacity for long-range modeling. Meanwhile, CNNs are widely recognized for their resource efficiency and translation invariance. Thus, an intuitive idea is to integrate the strengths of ViT to facilitate the evolution of CNNs. To this end, ConvNeXt~\cite{liu2022convnet} draws inspiration from ViT and incorporates effective long-distance modeling into its architecture, thereby outperforming ViT-based models in various vision tasks. The most critical insight is to enhance long-distance dependencies through the enlargement of kernel sizes.

\begin{figure}
    \begin{center}
            \includegraphics[width=0.99\linewidth]{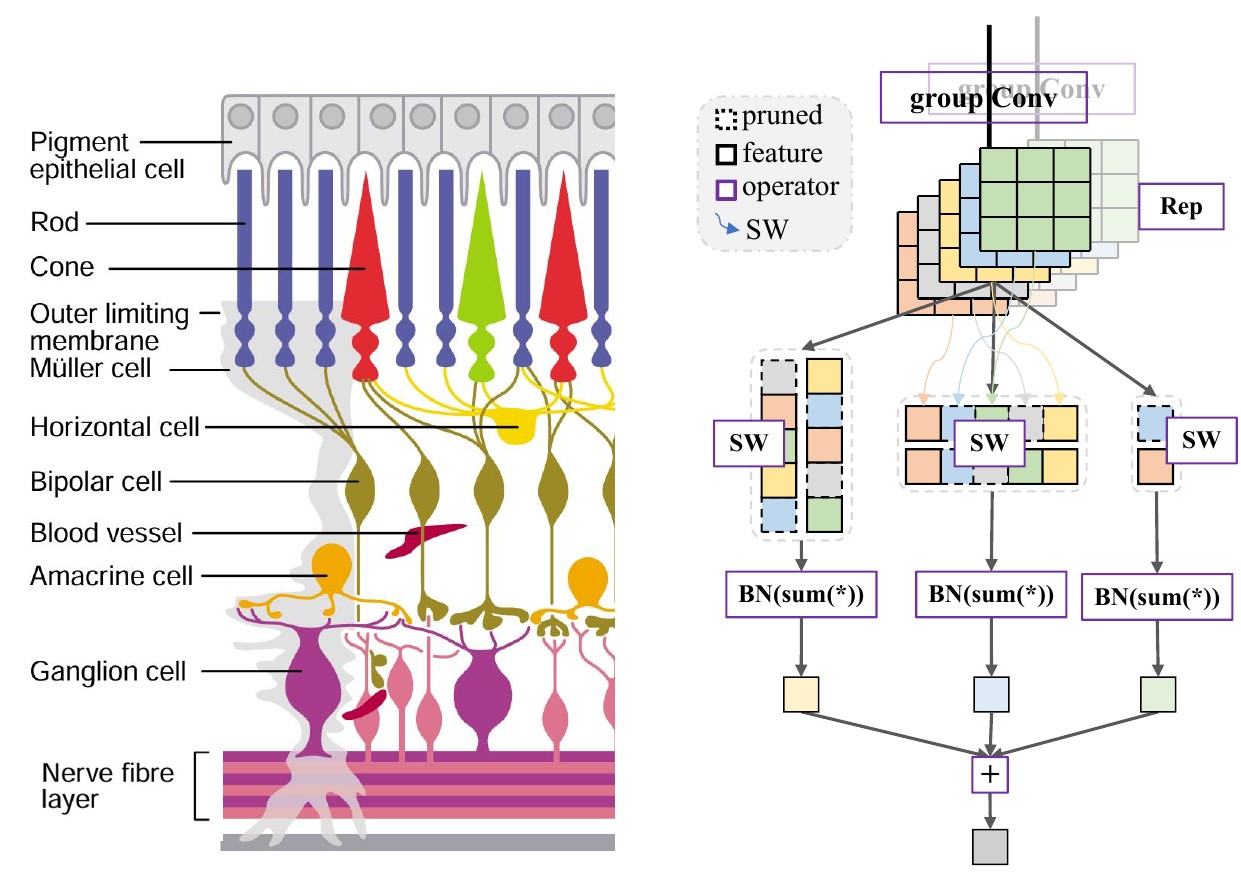}
        \vspace{-2mm}
        \caption{The left panel depicts the retina's cellular structure, which is composed of photoreceptor cells and ganglia, with photoreceptors sending visual signals to ganglia via multiple pathways.~~The right panel presents our Shiftwise (SW) Convolution, consisting of standard convolution and a connection-centric module. The approach utilizes group convolution and reparameterization (Rep) to extract basic information, which is then processed by a shift algorithm to mimic large kernel convolution.}
        \label{fig:f01}
        \vspace{-0.35in}
    \end{center}
\end{figure}

Motivate by the above insight, recent studies like RepLKNet~\cite{ding2022scaling}, SLaK~\cite{liu2022more}, and UniRepLKNet~\cite{ding2023unireplknet} have significantly advanced kernel size, and made CNNs great again over ViTs. More specifically, RepLKNet~\cite{ding2022scaling} expanded convolutional kernels from $7\!\times\!7$ in ConvNeXt~\cite{liu2022convnet} to $31\!\times\!31$, which led to performance improvements. SLaK~\cite{liu2022more} introduced decomposable convolutions, utilizing two distinct strip convolutions of $51\!\times\!5$ and $5\!\times\!51$, further advancing the capabilities of pure CNNs. ParCNetV2~\cite{xu2023parcnetv2} expanded kernel sizes to twice the feature map size, achieving accuracy levels nearly on par with SLaK~\cite{liu2022more}. In contrast, UniRepLKNet~\cite{ding2023unireplknet} opted for a comparatively smaller $13\!\times\!13$ kernel as the central component of its model, which significantly enhanced performance. Notably, existing works reveal that there's a point of diminishing returns in precision by simply enlarging the kernel size. 
Then, one must propose many tailored tricks to avoid diminishing returns in order to potentially enhance accuracy further.

The above phenomenon indicates that simply enlarging kernel sizes to achieve global attention is not an ideal solution for capturing long-distance dependencies. Meanwhile, large convolution kernels in CNNs have been phased out in favor of smaller ones, with long-range dependencies being established through multi-layer stacking \cite{simonyan2014very, guo2023visual}.
Thus, the question arises: Can long-distance dependency be effectively modeled through alternative mechanisms rather than just enlarging the convolution kernel, as has been the traditional evolution? Unlike the single-module approach to achieving large receptive fields, the human retina processes visual information through a network of interconnected cells. Specifically, cone and rod cells connect to ganglion cells via bipolar and horizontal cells, which then relay the visual signals to the brain, as depicted in the cellular organisation of the retina~\cite{WilkinsonBerka2004DiabetesAR}( Fig. ~\ref{fig:f01}). Before reaching the ganglion cells, photoreceptor cells are integrated in a variety of interconnected combinations. Similarly, CNNs can also have decoupled constructions by extracting basic signals at certain granular level and establishing long-distance dependencies through diverse connections.

However, the employment of large convolutional kernels has become nearly indispensable in modern CNNs. As studies~\cite{ding2023unireplknet, Luo2016UnderstandingTE} have shown, large kernel convolutions surpass the stacking of small convolutions in effectively expanding the effective receptive field (ERF). Consequently, when redesigning modules with a focus on decoupling long-distance dependencies, they should possess the following traits: i) Efficiency in broadening the receptive field. ii) Basic signal extraction with a fine-grained atomicity. iii) The ability to integrate features through diverse connections. Guided by these, in this paper, we propose Shiftwise (SW) Conv, a plug-and-play module that leverages standard $3\times3$ convolutions to achieve the effect of large kernels within a pure CNN framework. To enhance feature utilization, we employ a multi-channel feature shift fusion method. Additionally, the SW module can selectively eliminate certain filters to foster diverse interconnections without altering the overall network architecture.

From a novel decoupling perspective on long-range dependencies, our proposed method outperforms the performance of large kernel convolution models. Across diverse tasks—including ImageNet classification~\cite{5206848}, semantic segmentation on ADE20K~\cite{Zhou2016SemanticUO}, object detection/segmentation on COCO~\cite{Lin2014MicrosoftCC}, and monocular 3D object detection on nuScenes~\cite{caesar2020nuscenes}—our approach surpasses state-of-the-art large kernel convolutional models, such as Slak~\cite{liu2022more} and UniRepLKNet~\cite{ding2023unireplknet}. This reinforces the conclusion that small convolutions can effectively supplant large ones within the framework of modern CNNs, aligning with the insights of the VGG~\cite{simonyan2014very} model.

Our contributions can be summarized as three-fold:

\begin{itemize}
    \item[$\bullet$] A new scheme is proposed to effectively replace large convolutional kernels with small ones.
    \item[$\bullet$] A simple yet effective plug-and-play pure convolution operator is proposed to enhance CNN's performance.
    \item[$\bullet$] Our experimental results demonstrate that decoupling long-range dependencies is an effective strategy for improving CNN performance.
\end{itemize}

\section{Related Work}

\subsection{Large Convolutional Kernels}
Large convolutional kernels have been central to classic CNNs like AlexNet~\cite{krizhevsky2012imagenet}, ResNet~\cite{he2015deep}, and Inception V1~\cite{szegedy2014going}. VGG's advocacy for small kernel stacking influenced model efficiency and development for a long time~\cite{he2018bag}. 
The recent resurgence in interest in large kernels, sparked by the ViT's~\cite{Dosovitskiy2020AnII} capabilities, has led to a focus on two main areas: the development of innovative CNN-Transformer hybrids~\cite{9711272, 10040235} and the modernization of CNN structures like ConvNeXts~\cite{liu2022convnet}.

Following ConvNeXts, studies have investigated the correlation between convolutional kernel size and CNN performance. The Visual Attention Network (VAN) ~\cite{guo2023visual} uses decomposable convolutions combined with a multi-layer stacking strategy to mimic large kernel convolutions, thereby enhancing spatial perception and improving model performance. Works such as RepLKNet ~\cite{ding2022scaling} and SLaK ~\cite{liu2022more} expanded kernel sizes to $31\times31$ and $51\times51$, respectively, setting new performance benchmarks and inspiring our proposed method. ParCNetV2~\cite{xu2023parcnetv2} further increased kernel sizes to twice the input feature map size, achieving performance on par with SLaK. PeLK~\cite{10656330}, taking cues from human focal and blur mechanisms, scales CNN kernels to 101×101, markedly boosting performance. UniRepLKNet~\cite{ding2023unireplknet} incorporates a smaller $13\times13$ dilated convolution with the squeeze-and-excitation (SE)~\cite{8578843} module for enhanced feature fusion, reducing the parameter count while outperforming SLaK.

\subsection{Shift Operation}
The Shift operator, as introduced by Wu et al.~\cite{wu2017shift} in 2007, enables the generation of diverse convolutional outputs at a reduced computational cost. Building upon this foundation, Active Shift ~\cite{jeon2018constructing} and Sparse Shift ~\cite{8953442} have incorporated learnable parameters to enhance their functionality. ACmix~\cite{pan2022integration}, Shift Graph Convolutional Network~\cite{ShiftGCN}, Xvolution~\cite{chen2021xvolution} and ShiftViT~\cite{Wang2022WhenSO} have all employed the Shift operation to expand the receptive field, thereby enabling more comprehensive attention mechanisms. 
These methods view the shift operator as a form of specialized Depthwise Convolution. 
Our approach goes beyond these by proposing a method to stack convolutions spatially, effectively emulating the impact of large kernel convolutions. Additionally, we employ a data-driven approach to extract the spatial structure of sparse large kernel convolutions. While AKConv ~\cite{zhang2023akconv} and Dilated Convolution with Learnable Spacings (DCLS)~\cite{khalfaoui2021dilated} feature offset operations akin to dynamic convolutions. In contrast, our method emphasizes the certain granularity of information extraction rather than pursuing the finer-grained manipulations.

\subsection{Ghost and Re-parameterization}
The $ghost$ concept, introduced by GhostNet~\cite{han2020ghostnet}, involves selectively processing features within each module and combining them with the rest for output. This idea has been further developed in networks like CSPNet~\cite{9150780} and FasterNet~\cite{10203371}. Re-parameterization techniques, as seen in RepVGG~\cite{ding2021repvgg}, enhance model complexity during training and simplify during inference for speed. ExpandNets ~\cite{guo2021expandnets}, MobileOne ~\cite{vasu2023mobileone} and VanillaNet ~\cite{chen2023vanillanet} have expanded on this by decomposing weight matrices and adjusting model width and depth. RepLKNet ~\cite{ding2022scaling}, SLaK-net ~\cite{liu2022more} and UniRepLKNet ~\cite{ding2023unireplknet} also utilize re-parameterization, showcasing its broad applicability in network design.

\section{Architecture Design}
\label{sec:architecture}

\subsection{Background} 

\begin{figure}
	\begin{center}
		\includegraphics[width=0.99\linewidth]{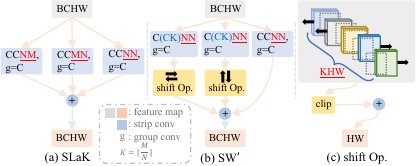}
		\vspace{-2mm}
		\caption{(a) SLaK's large kernel convolution architecture employs a decomposable separable convolution approach using $M \times N$ and $N \times M$ strip convolutions. (b) One-to-many separable convolution with proper feature movement can be equivalent to SLaK's strip convolution. (c) Overview of the proposed shift operation. The addition operation preserves the network's structural integrity even after coarse-grained pruning.
        }
		\label{fig:Replacement}
		\vspace{-0.35in}
	\end{center}
\end{figure}

Aiming to emulate the traditional evolution triggered by VGG—where large-kernel convolutions are replaced by small convolutions stacked in depth—we discovered and validated a method to stack receptive fields in width. SLaK enhances both kernel size and model performance to new levels, thus serving as the foundation for our research.
It employs a ResNet variant with $M \times N$ strip convolutions across four stages, where $M$ is [51, 49, 47, 13] and $N$ is consistently 5. The large-kernel convolution uses two parallel strip convolution branches, complemented by a standard small convolution, as shown in Fig. \ref{fig:Replacement} (a).
Additionally, SLaK employs a width increase factor ``$r$'' to boost model performance. To counterbalance the rise in model parameters, SLaK incorporates a fine-grained sparsity ``$s$'' that converts the weights of convolution and Feed Forward Network (FFN) into sparse matrices, with default values of $r$=1.3 and $s$=0.4. To track the scaling trends of large kernel sizes, SLaK models are initially trained for 120 epochs and then extended to 300 epochs to align with state-of-the-art models. We've adopted this training approach in our subsequent experiments. Assuming feature map dimensions are $H\! \times\! W$ and input channels are $C$.

In this section, we explore key aspects of our architecture using a structured approach akin to ConvNeXt. We evolve the initial SW module to match the SLaK strip convolution (§\ref{sec_equ}), eliminate redundant connections (§\ref{sec_redu}), and counter SLaK's width-expanding design (§\ref{sec_width}). Additionally, we investigate multi-path feature fusion (§\ref{sec_random}), localized reparameterization adaptation (§\ref{sec_multi_rep&path}), localized pruning adaptation (§\ref{pruning}), backbone optimization (§\ref{unirep}), and overall architectural design (§\ref{sec:Specifications}). Collectively, these efforts aim to merge LK into the mainstream of CNN research.

\subsection{Replacement Experiment}
\label{sec_equ}

In our initial modification, we replaced the $M \times N$ convolutions employed in SLaK (denoted as \#0) with a series of standard $N \times N$ convolutions, as illustrated in Fig. \ref{fig:Replacement} (b). 
This was achieved by employing group convolutions with the number of groups matching the input channels $C$, and the output channels were set to $\left \lceil \frac{M}{N} \right \rceil$, where $\left \lceil \right \rceil$ denotes the ceiling function. Subsequently, we applied appropriate offsets to the output features based on their sequence numbers to ensure that the outputs align perfectly with the offset positions of the original $M \times N$ convolution when laid out (Fig. \ref{fig:Replacement} (c)). For this replacement experiment (the laid out version labeled as \#1 for infer\&train) , we should adjusted the padding ($\delta_{p}=P_{real}-N//2$) and offset of the outputs of standard convolutions to account for the discrepancies in initial and final sliding window positions due to the different kernel sizes. The padding may cause a slight adjustment in the parameters, resulting in the height ${H}'$ being $H+\delta_{p}$, and the width ${W}'$ similarly. 
Employing SLaK-trained model parameters, this spatial stacking approach maintained accuracy, achieving 82.5\% during inference.

We proceeded to train the modified version of the model. Instead of employing the fine-grained pruning method utilized by SLaK, we opted for a coarse-grained pruning strategy that remove filters during training. Because based on the existing general hardware, the coarse-grained pruning can reduce a lot of computation. 
This was achieved by modifying the pruning method used in SLaK to a coarse-grained approach,
where we summed the absolute values of all parameters within a filter and pruned those with lower rankings. For more information, see the parameter prune-and-grow strategy as detailed in ~\cite{Liu2021DoWA, liu2022more}. Given that the modified version has multiple output channels per input channel, it is unlikely that all output channels will be pruned. This ensures that the final structure of the module remains intact, which is typically an attribute of fine-grained pruning. The sparse version, labeled as \#1, reached an accuracy of 82.27\% at 300 epochs, showing minimal decline.

\begin{table}[t!]
	\centering
	\renewcommand\arraystretch{0.89}
	\setlength{\tabcolsep}{0.9mm}
	\footnotesize
	\caption{\textbf{Kernel Replacement Experiment}. 
    Experiments on progressively replacing large kernels with small ones based on SLaK.
    $HW$ represents the feature map dimensions, with $C$ indicating the number of channels. Different padding configurations result in variations in parameter calculations, denoted as ${H}'\!{W}'$ and ${H}''\!{W}''$. The ceiling function is indicated by $\left \lceil \right \rceil$. Experiments were conducted for 120 epochs to outline the scaling trends while reducing the training load. When the kernel size is modified, denoted as  $N_{*}$, it implies ${N=*}$. Parameter count calculations mainly focus on two strip convolutions here.}
	\vspace{-0.1in}
	\begin{tabular}{l|l|c|c|l}
\hline
	& \multirow{2}{*}{Method}&  IN-1K acc. & IN-1K acc. & \multirow{2}{*}{ \#param.} \\
       & &(300epoch) &(120epoch)& \\ 
    
    \hline
	\#0 & Slak tiny   	  	   & 82.5  & 81.6      & \tiny$H\!W\!M\!N \times 2$ \\
	\#1 & laid out(infer)	   & 82.5  & -     	   & \tiny ${H}'\!{W}'\!\left\lceil\!\frac{M}{N}\!\right\rceil \!N\!N \times 2$   \\
	\#1 & laid out(train) 	   & 82.27 & -    	   & \tiny As above   \\
	\hline
	\#2 & SW-(pad=N-1)         & 82.38 & 81.36   & \tiny ${H}''\!{W}''\!\left\lceil\!\frac{M}{N}\!\right\rceil \!N\!N$  \\
	\#3 & SW-(pad=N//2)        & - 	 & 81.26     & \tiny $H\!W\!\left\lceil\!\frac{M}{N}\!\right\rceil \!N\!N$  \\
	\#4 & Rep*2                & - 	 & 81.52     & \tiny As above   \\
	\#5 & Relocate BN          & - 	 & \textbf{81.65} & \tiny $\#4+(C\!\times\! 2)\!\times\! 3 $ \\
    \hline
	\#6 &  $G$=0.23            & - 	 & 81.34     & \tiny $\#5\times (1-G) $ \\
        \#7 &  $N$=5 to $N$=3      & - 	 & 81.44     & \tiny $(H\!W\!\left\lceil\!\frac{M}{N_{3}}\!\right\rceil \!N_{3}\!N_{3})\times (1-G)$         \\
        \#8 &  Rep*4               & - 	 & \textbf{81.60} & \tiny As above   \\ 
    \hline
        
\end{tabular}
	\vspace{-0.15in}
	\label{table-t31_32}
\end{table}

\subsection{Eliminate Redundant Connections}
\label{sec_redu}

\begin{figure}
	\begin{center}
		\includegraphics[width=0.69\linewidth]{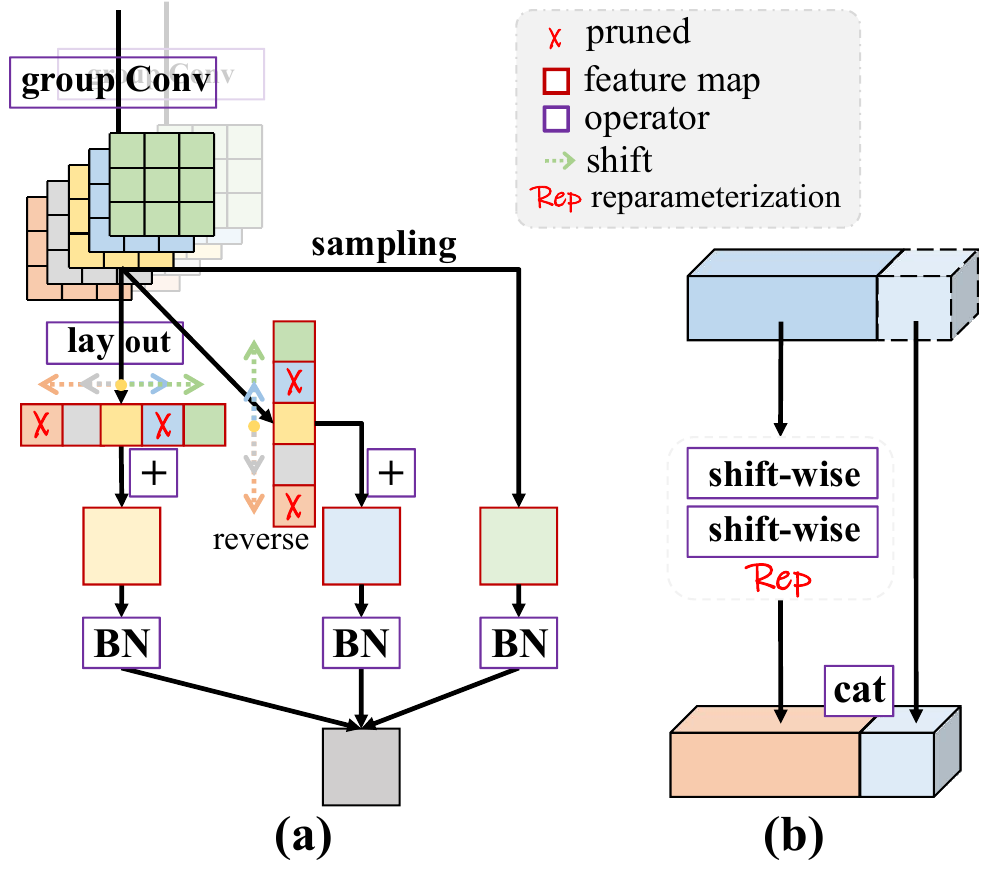}
		\vspace{-2mm}
		\caption{(a) 
        Fig. \ref{fig:Replacement} shows that the similar structures of the convolutional branches lead to minimal variation across the data manifold. This implies that the branches can be consolidated.
        The shift operation introduces more divergence in the data manifold, necessitating the repositioning of the Batch Normalization (BN) layer to follow the shift operation for optimal performance.
		(b) Reducing network width with a ghost-like approach to counterbalance SLaK's expansion.}
		\label{fig:connections}
		\vspace{-0.35in}
	\end{center}
\end{figure}

Based on this, we explored new configurations: i) Recognizing the similarity between the two decomposing branches originating from SLaK's strip convolution, we restructured them to utilize a shared convolution output. To avoid correlation, the second branch was modified to employ offsets in reverse order. As illustrated in Fig. \ref{fig:connections} (a). ii) The padding, influenced by diverse feature sizes and large kernels size, was deemed suboptimal for network design. To address this, we standardized the padding values. Implementing these configurations, we set the padding values to $N\!-\!1$ (labeled as \#2) and $N//2$ (labeled as \#3, default). In this scenario, the total number of parameters is halved compared to SLaK's, with $H'' =H\!+ \!((N\! -\! 1)\! -\! \left \lceil\! \frac{N}{2}\! \right \rceil)$. After 120 epoch training, the accuracy of the model is almost unchanged, which is 81.36\% and 81.26\%, respectively.

To counteract the performance decline, we implemented the popular re-parameterization technique (Rep). We integrated two Rep branches and applied a shared mask to both within our module. This approach harnesses the advantages of both Rep and sparsity (labeled as \#4). The model achieved an accuracy of 81.52\%, which is almost close to the accuracy of SLaK at 120 epoch. In SLaK's strip convolution, each convolution is followed by a batch normalization (BN) for each branch. However, the data manifolds from multiple Rep branches tend to be similar, resulting in a low degree of difference, whereas the outputs from SW branches exhibit a high degree of difference due to their varied laid out directions. Therefore, we removed the BN after the convolution and added one after the SW branches (labeled as \#5). Because of the BN layer, Experiment \#5 has $3C \times 2$ more parameters in the large kernel convolution branch than Experiment \#4. 
From experiment \#0 to \#5, we have successfully used small convolution over SLaK large kernel convolution after such a series of changes. At the same time, the number of parameters involved in the construction of large kernel convolution is almost halved. These experiments also demonstrate the efficacy of decoupling long-distance dependency modeling into two distinct parts: feature extraction at a fundamental atomic granularity and long-distance modeling that emphasizes connectivity as the primary characteristic.

\subsection{Reduce Parameters}
\label{sec_width}

We must admit that, like the $M \times N$ large convolution kernel used by SLaK, only the convolution center point can slide the whole map during the sliding window process. For other $N \times N$ windows, such as the small convolution at the end side of the $M \times N$ large one, a considerable proportion of the sliding positions are in the padding 0 value region. This is also an important place for programming optimization of large kernel convolution operator.
At the same time, expanding the network width is essential for SLaK's final performance, a feature that our current structure \#5 retains. 
We must reduce parameters to mitigate the impact of network expansion and the computational load from adjusting  kernels.
We use a slimming technique that allows only a subset of channels to contribute to the large convolution kernel. This is achieved through a ghost-like approach, setting a ratio $G$ that specifies the fraction of channels to pass directly to the subsequent layer. By slimming the large kernel, we can decrease the number of parameters while keeping the input channels of the large kernel constant with expansion ratio $R$, ensuring that $G$ satisfies the equation $R(1-G)=1$. For example, if SLaK uses a width ratio $R$ of 1.3, we set $G$ to 0.23 to offset the effects of increased width on large convolutions (\#6), as shown in Fig. \ref{fig:connections} (b). With this adjustment, we further reduce the convolution kernel size to a smaller $3\times3$ convolution (\#7).

Compare models \#6 and \#7, which, while initially aimed at reducing the total parameter volume with minimal performance impact, surprisingly benefited from selecting a smaller kernel size. This indicates that an appropriate level of granularity is beneficial to long-distance dependence. 
Additionally, we tested increasing the number of parallel Rep blocks (\#8), which resulted in an accuracy of 81.60\%, matching SLaK's performance. In the remainder of this paper, we set $ N = 3 $ as default.

\subsection{Enhance Utilization}
\label{sec_random}

\begin{figure}
	\begin{center}
		\includegraphics[width=0.99\linewidth]{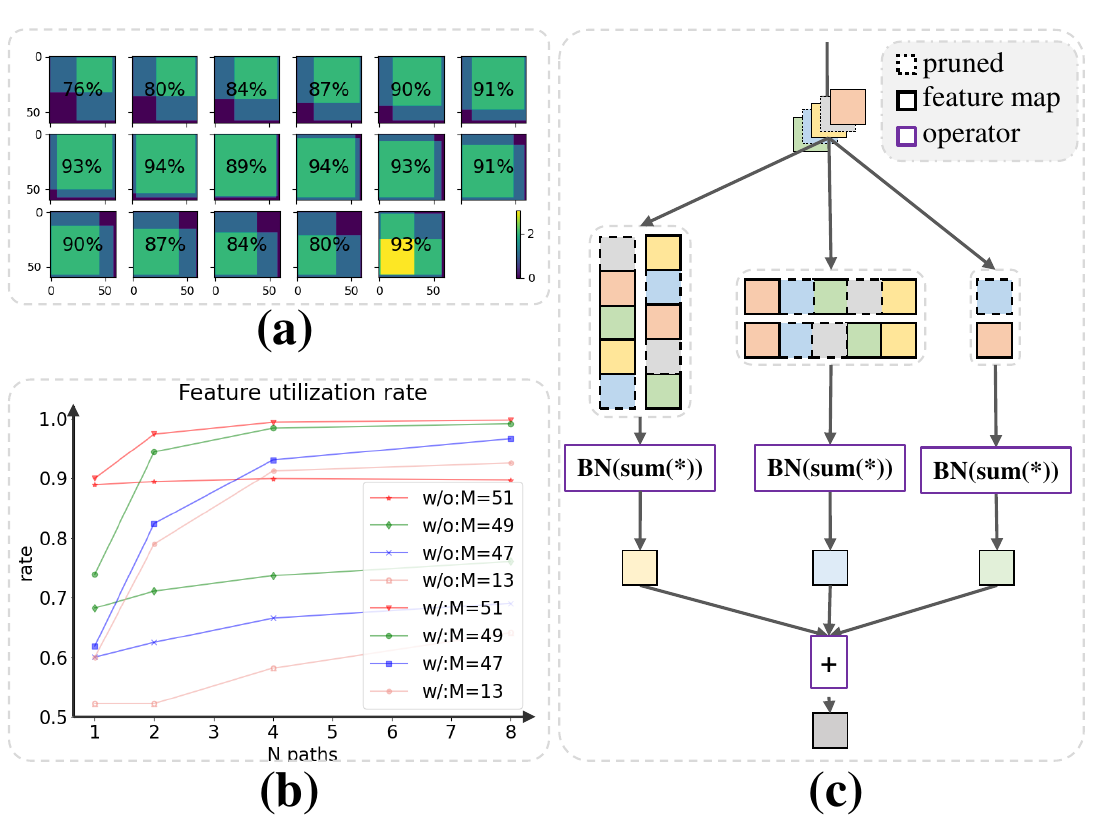}
		\vspace{-2mm}
		\caption{(a) When the equivalent large kernel convolution size is $51\!\times\! 3$ , the coverage regions of the feature maps, along with their utilization proportions, are determined by the areas that can be propagated downward through the shift operation. (b)The utilization ratio of feature maps varies as the number of downward propagation paths increases. $M$ denotes the longer side of the equivalent large kernel convolution, while $w/$ and $w/o$ indicate whether shuffled ordering is applied during the multi-path propagation process. (c) The network structure with increased downward propagation paths.}
		\label{fig:multiedge}
		\vspace{-0.35in}
	\end{center}
\end{figure}

When using SW operation on group convolution, we found that the utilization ratio of feature maps is low. Specifically, we label the feature maps processed by the three branches of SW to obtain an image of the coverage distribution. As shown in Fig. \ref{fig:multiedge} (a), for specific feature maps, the coverage is predictable and the utilization ratio is limited. If we refer to the three branches of SW as one edge, denoted as $E$, as the edges increases, the given order of combined feature maps does not improve the coverage. When we choose to combine feature maps in a random order for each edge, the coverage is greatly improved as the number of edges increases, as shown in Fig. \ref{fig:multiedge} b.
To assess the impact of feature utilization, we based our analysis on Experiment \#3. To break predictability, we introduced disordered offset calculations in one branch (\#9). To boost feature map utilization, we added additional edges for channel order mapping in our experiments (\#10, Fig. \ref{fig:multiedge} c). As shown in Table \ref{tab:Combining}, performance consistently rose to 81.36\% and 81.51\%.
Because the data read and written here is stored in continuous locations, it can minimize the impact on algorithm efficiency. Please refer to Appendix~\cref{sec:algorithm} for more details.

\subsection{Integrating Multiple Rep and Edges}
\label{sec_multi_rep&path}

Considering the significant improvement in performance of multiple branches of Rep and edges, we attempted to merge the two experiments \#8 and \#10, labeled as \#11. We found that it significantly outperformed SLaK, reaching an accuracy of 81.82\% with a setup featuring 2 Rep branches and 4 SW edges. 
We therefore continued to experiment with different numbers branches of Rep and edges to see how their configurations can bring benefits, as shown in experiments \#12 to \#14.
However, their final performance differences are almost negligible, which marks a sharp contrast with experiments \#4 and \#8. This discovery seems to indicate an overlapping effect between reparameterization and the reuse of feature maps.
Upon in-depth analysis, we suspected that the initial sparsity distribution across layers might be the primary cause. Consequently, we experimented with using initial sparsity as a variable. There are many ways to affect the initial sparsity, and we adopt following approach here. Instead of adding all block weights first and then calculating sparsity, we calculated the sparsity of each Rep branches and then averaged these values within the same layer to determine the initial sparsity. This experiment, labeled \#15, achieved the highest accuracy of 81.94\%. We also tried reducing overall sparsity to directly affect initial sparsity, which yielded similar results in experiment \#16, matching \#15.

At this moment, the SW architecture has largely taken shape, as shown in Fig. \ref{fig:f01}. The $3 \times 3$ standard convolutions are tasked with the appropriate granularity for capturing fundamental visual details. 
The establishment of long-distance dependencies is delegated to the SW module, a method that relies on connections. These two parts effectively decouple the large kernel convolution and improve the performance of the model.

\begin{table}[t!]
	\centering
	\renewcommand\arraystretch{0.89}
	\setlength{\tabcolsep}{0.9mm}
	\footnotesize
	\caption{\textbf{Experiments on Enhancing Feature Utilization}. For detailed information, refer to sections 3.5 to 3.8.}
	\vspace{-0.1in}
	\begin{tabular}{l|l|c}
    \hline
            &          model              & IN-1K acc.(120epoch) \\
    \hline
    \#3   & SW-(pad=N//2)  	              & 81.26      \\
    \#9   & disordered offset             & 81.36      \\
    \#10  & two edges		              & 81.51      \\
    \hline
    \#11  & rep2 E4                       & 81.82     \\
    \#12  & rep4 E4                       & 81.79     \\
    \#13  & rep4 E2                       & 81.77     \\
    \#14  & rep8 E4                       & 81.85     \\
    \#15  & rep2 E4 mean                  & \textbf{81.94}    \\
    \#16  & rep2 E4 sparsity 0.4 to 0.3   & 81.93     \\
    \hline
    \#17  & \#16+gap=3                    & \textbf{81.95}    \\
    \#18  & \#16+gap=5                    & 81.94  \\
    \hline
    \#19  & \ architecture                & 82.25 \\
    \#20  &    +SE                        & \textbf{82.27} \\
    \#21  &   stage 1                     & 82.17   \\
    \#22  & subset                        & 81.72  \\
    \#23  & interleaving                  & 82.23   \\
    \hline  
\end{tabular}
	\vspace{-0.15in}
	\label{tab:Combining}
\end{table}

\subsection{Shared Mask}
\label{pruning}
Inspired by Experiment \#15, we focused on hyperparameter settings related to sparsity. A natural question arises: Do different Rep branches in the same layer need to maintain consistent sparse masks throughout the training process?
To explore this, we reduced the frequency of mask sharing among branches, as demonstrated in experiments \#17 and \#18.
By reducing the frequency of sparse mask sharing between branches, the performance of the model can be effectively improved. This indicates that there are differences in feature extraction between Rep branches. Reducing the synchronization frequency of masks can allow the model to explore different combinations of filters.

\subsection{Infrastructure}
\label{unirep}
We also note a significant architectural differences between UniRepLKNet and SLaK. UniRepLKNet favors increasing depth to boost representational capacity, whereas SLaK aims to ``use more groups, expand width.'' This recalls discussions around VGG's design philosophy. 
To verify which idea of depth first or width first can bring more performance benefits to our proposed operator in modern CNNs, 
we conducted an experiment using UniRepLKNet's architectural hyperparameters only, including block counts in four stages and channel numbers per stage, labeled as \#19. This experiment achieved an accuracy of \textbf{82.25}\%. 
Once again, it has been confirmed that under the constraint of the total number of parameters, deepening the network provides better model performance than widening the network.
When training with 120 epochs on ImageNet 1k, networks were constructed based on the hyperparameters of SLaK and UniRepLKNet, respectively. Our proposed operator exceeded SLaK's 81.6\% and UniRepLKNet's 81.6\%, as shown in experiments \#5 and \#19.

Recognizing the substantial benefits of UniRepLKNet's network architecture, we explored additional parameters from this model. Firstly, we incorporated the SE module, which UniRepLKNet highlights, in our experiment labeled as \#20, achieving an accuracy of 82.27\%. Secondly, we experimented with eliminating large convolution kernels in the first stage, labeled as \#21, resulting in an accuracy of 82.17\%. The reparameterized branches in UniRepLKNet use multiple dilated convolutions, and their dilation rates appear as sequences with increasing sparsity. Following this method, we randomly generate subsets of masks for multiple Rep branches when sharing sparse masks(\#22). But the accuracy has decreased to 81.72\%. Another key feature of UniRepLKNet is the interleaved use of small kernel and large kernel in stage 3 to reduce the number of parameters.
In stages 2 to 4, UniRepLKNet utilizes a convolution kernel size of $13 \times 13$, which corresponds to our parameter count of approximately [$49\!\times\!3$, $47\!\times\!3$, $13\!\times\!3$], representing [0.87, 0.83, 0.23] times the count of the previous stage. 
Although our module does not consume an excessive number of parameters,
we also explore the benefits of incorporating different types of convolution kernels (\#23). The overall accuracy has slightly decreased to 82.23\%.
Apart from the network architecture and the Squeeze-and-Excitation (SE) mechanism, other enhancements present in UniRepLKNet cannot be directly integrated into our proposed algorithm to yield performance improvements.

\subsection{Architectural Specifications}
\label{sec:Specifications}
Thus far, we've identified a strategy that effectively replaces large kernels with small $3\!\times\!3$ standard convolutions, significantly enhancing performance. This approach revealed several parameters crucial for refining our architecture. 

The architecture begins with two stride-2 $3\!\times\!3$ convolutional layers in the stem, converting raw input into $C$ channel feature maps, where $C$ is an architectural hyper-parameter. Subsequent downsampling blocks use one stride-2 $3\!\times\!3$ convolutional layer each, doubling the channel count, leading to channel configurations of $C$, $2C$, $4C$, and $8C$ across the four stages. The block counts per stage are [3, 3, 18, 3] for SW-tiny and [3, 3, 27, 3] for SW-small. The initial stage channels $C$ are set to 80 for SW-tiny and 96 for SW-small, followed UniRepLKNet's hyperparameters. The equivalent large convolution sizes $M$ for each stage, inherited from SLaK, are [51, 49, 47, 13], meaning our group convolution's output expansion factor is $\left\lceil\!\frac{M}{3}\!\right\rceil$. For instance, in the first stage with $C$ channels, the group convolution outputs $C\!\times\!\left\lceil\!\frac{M}{3}\!\right\rceil$ channels, which are then processed by the SW operator, as shown in Fig. \ref{fig:f01}.


\section{Results}
\label{sec:Results}
We employ SW to benchmark against state-of-the-art models on ImageNet classification~\cite{5206848}, semantic segmentation on ADE20K~\cite{Zhou2016SemanticUO},   object detection/segmentation on COCO~\cite{Lin2014MicrosoftCC}, and monocular 3D object detection on nuScenes~\cite{caesar2020nuscenes}.

\begin{table}[t!]
	\centering
	\renewcommand\arraystretch{0.89}
	\setlength{\tabcolsep}{0.9mm}
	\footnotesize
	\caption{\textbf{ImageNet classification}. For detailed information on Throughput, please refer to Appendix ~\cref{sec:throughout}. ``T/C'' denote transformer/ConvNet.}
	\vspace{-0.1in}
	\begin{tabular}{l|c|c|c|c|c}
\hline
    \multirow{2}{*}{Method} & \multirow{2}{*}{Type} & Input & Params & FLOPs & Acc \\
    & & size &(M)&(G)&(\%) \\ 

    \hline
    \rowcolor{gray!20}
    \textbf{SW-tiny}                            &   C   &   $224^2$     &   31      &   5.0     &   \textbf{83.4}\\
    UniRepLKNet-T\cite{ding2023unireplknet}     &   C   &   $224^2$     &   31      &   4.9     &   83.2    \\
    FastViT-SA24~\cite{vasu2023fastvit}         &   T   &   $256^2$     &   21      &   3.8     &   82.6    \\
    PVTv2-B2~\cite{wang2022pvt}                 & T     & $224^2$       & 25        & 4.0       & 82.0 \\
    CoAtNet-0~\cite{dai2021coatnet}             & T     &$224^2$        & 25        & 4.2       & 81.6   \\
    DeiT III-S~\cite{touvron2022deit}           & T     &$224^2$        & 22        & 4.6       &   81.4\\
    SwinV2-T/8~\cite{liu2022swin}               & T     &$256^2$        & 28        & 6         & 81.8 \\
    SLaK-T~\cite{liu2022more}                   & C     & $224^2$       & 30        & 5.0       &82.5  \\
    \hline
    \rowcolor{gray!20}
    \textbf{SW-small}                           &   C   &   $224^2$     &   56      &   9.4     &   \textbf{83.9}    \\
    UniRepLKNet-S\cite{ding2023unireplknet}     &   C   &   $224^2$     &   56      &   9.1     &    83.9    \\
    ConvNeXt-S~\cite{liu2022convnet}            &   C   &   $224^2$     &   50      &   8.7     &   83.1\\
    HorNet-T~\cite{rao2022hornet}               & C     & $224^2$       & 23        & 3.9       &   83.0 \\
    FastViT-SA36~\cite{vasu2023fastvit}         &   T   &   $256^2$     &   30      &   5.6     &   83.6     \\
    CoAtNet-1~\cite{dai2021coatnet}             & T     &$224^2$        & 42        & 8.4       & 83.3  \\
    SLaK-S~\cite{liu2022more}                   & C     &$224^2$        & 55        & 9.8       &83.8 \\
    FastViT-MA36~\cite{vasu2023fastvit}         &   T   &   $256^2$     &   43      &   7.9     &   83.9    \\
    SwinV2-S/8~\cite{liu2022swin}               & T     &$256^2$        & 50        & 12        & 83.7\\
    RepLKNet-31B~\cite{ding2022scaling}         & C     &$224^2$        & 79        & 15.3      & 83.5\\
    PVTv2-B5~\cite{wang2022pvt}                 & T     & $224^2$       & 82        & 11.8      & 83.8 \\
    
    \hline

\end{tabular}
	\vspace{-0.15in}
	\label{table:imgnet}
\end{table}

\noindent\textbf{Imagenet-1k classification}.
We employ the identical training settings as described in Section ~\cref{sec:Specifications}, now extending the training to the complete 300 epochs. 
Models of different sizes need specific hyperparameters, performance at 120 epochs is indicative but not definitive; thus, extended training at 300 epochs is crucial for accurate hyperparameter tuning.
The addition of new parameter settings increases the number of variables to explore, thus boosting resource consumption for larger models.
Noting that UniRepLKNet did not experiment with the base model on ImageNet-1K, we focus on presenting the tiny and small model results here. For further details, please refer to Appendix ~\cref{sec:scalinglaw}.

As shown in Table~\ref{table:imgnet}, we compared the performance of SW on ImageNet-1K against various state-of-the-art models. With comparable model sizes and FLOPs, our approach outperforms both SLaK and UniRepLKNet, two of the top-performing pure CNN models. Utilizing small convolutions, our method achieves higher accuracy than state-of-the-art transformers, such as Swin Transformer~\cite{liu2022swin} and FastViT~\cite{vasu2023fastvit}. This demonstrates the enhanced efficacy of pure convolutional networks, showing that standard $3 \times 3$ convolutions can match the effects of larger convolutions and further boost the performance of CNNs.

\noindent\textbf{COCO object detection and instance segmentation}.
We utilize the pretrained SW models as backbones for Cascade Mask R-CNN~\cite{he2017mask,cai2019cascade} and follow the standard 3x (36-epoch) training recipe with MMDetection~\cite{mmdetection}. As shown in Table~\ref{tab:det}, SW exceeds the performance of UniRepLKNets, ConvNeXt, RepLKNet, and SLaK—key examples of Vision Transformers and established large-kernel ConvNets.

\begin{table}[t]
        \centering
    \renewcommand\arraystretch{0.89}
    \setlength{\tabcolsep}{0.9mm}
    \footnotesize
    \caption{\textbf{Object detection on COCO validation set}. The FLOPs were calculated using inputs of 800 by 1280 pixels.}
    \vspace{-0.1in}
    
\begin{tabular}{l|c|c|c|c}
\hline
    Method & Params (M) &  FLOPs (G)  & $\text{AP}^{\text{box}}$  &  $\text{AP}^{\text{mask}}$   \\
    \hline
    \rowcolor{gray!20}
    \textbf{SW-tiny}   &          87  &   751     &   \textbf{52.21}    &   \textbf{45.19}     \\
    UniRepLKNet-T~\cite{ding2023unireplknet}   &          89  &   749     &   51.8    &   44.9     \\
    Swin-T~\cite{liu2021swin}      &   86   &   745     &   50.4    &   43.7\\
    ConvNeXt-T~\cite{liu2022convnet}  &   86   &   741      &   50.4    &   43.7\\
    SLaK-T~\cite{liu2022more}      &   -   &   -        &   51.3    &   44.3\\
    
    \hline
    \rowcolor{gray!20}
    \textbf{SW-small}       &   110  &   839    &   52.7     &    45.67    \\ 
    UniRepLKNet-S ~\cite{ding2023unireplknet}      &   113  &   835    &   53.0      &     45.9      \\ 
    Swin-S~\cite{liu2021swin}      &   107   &   838    &   51.9    &   45.0\\
    ConvNeXt-S~\cite{liu2022convnet}  &   108   &   827     &   51.9    &   45.0\\
    SLaK-T~\cite{liu2022more}      &   -   &   -        &   51.3    &   44.3\\
    
    \hline

\end{tabular}
    \label{tab:det}
    \vspace{-0.05in}
\end{table}

\noindent\textbf{ADE20K semantic segmentation}. 
We integrate pretrained SW models into UPerNet~\cite{xiao2018unified} for ADE20K~\cite{Zhou2016SemanticUO} segmentation tasks, adhering to the standard 160k-iteration training protocol from MMSegmentation. The table~\ref{tab:seg} presents the mean Intersection over Union (mIoU) scores on the validation set. Our method holds significant advantages over other models.

\begin{table}[t]
    \centering
    \setlength{\tabcolsep}{1.3mm}
    \footnotesize
    \caption{\textbf{Semantic segmentation on ADE20K validation set}. The FLOPs were calculated using inputs of 512 by 2048 pixels. $SS$ and $MS$ denote single-scale and multi-scale evaluations, respectively.}
    \vspace{-0.1in}
    \begin{tabular}{l|c|c|c|cc}
    \hline
    \multirow{2}{*}{Method} & Crop & Params & FLOPs & mIoU & mIoU\\
    	& size & (M)&(G) & (SS) & (MS)   \\
    \hline 
    \rowcolor{gray!20}
        \textbf{SW-tiny}                            &   512$^2$     & 62    & 948   & \textbf{49.22}    &   \textbf{50.06}   \\ 
        ConvNeXt-T~\cite{liu2022convnet}            & 512$^2$       & 60    & 939   & 46.0              & 46.7 \\
        SLaK-T~\cite{liu2022more}                   & 512$^2$       & 65    & 936   & 47.6              & - \\
        UniRepLKNet-T~\cite{ding2023unireplknet}    &   512$^2$     & 61    & 946   &   48.6            &   49.1   \\ 
        Swin-T~\cite{liu2021swin}                   & 512$^2$       & 60    & 945   & 44.5              & 45.8 \\
        InternImage-T~\cite{wang2023internimage}    & 512$^2$       & 59    & 944   & 47.9              & 48.1  \\
     \rowcolor{gray!20}
     
    	\hline
     \rowcolor{gray!20}
        \textbf{SW-small}                           &   512$^2$     & 88    & 1039      & 49.79     & 50.83 \\
        ConvNeXt-S~\cite{liu2022convnet}            & 512$^2$       & 82    & 1027      & 48.7      & 49.6  \\
        SLaK-S~\cite{liu2022more}                   & 512$^2$       & 91    & 1028      & 49.4      & - \\
        UniRepLKNet-S~\cite{ding2023unireplknet}    &   512$^2$     & 86    & 1036      & 50.5      & 51.0  \\
        Swin-S~\cite{liu2021swin}                   & 512$^2$       & 81    & 1038      & 47.6      & 49.5 \\
        InternImage-S~\cite{wang2023internimage}    & 512$^2$       & 80    & 1017      & 50.1      & 50.9 \\
        \hline

    \end{tabular}
    \label{tab:seg}
    \vspace{-0.1in}
\end{table}

\noindent\textbf{Monocular 3D object detection}. 
We deploy pretrained SW models as the foundation for fcos3d~\cite{wang2021fcos3d} on the nuScenes~\cite{caesar2020nuscenes} dataset, utilizing MMDetection3D~\cite{mmdet3d2020} for our setup. The table~\ref{tab:seg} details the mean Average Precision (mAP) obtained on the validation set.

\begin{table}[t]
        \centering
    \renewcommand\arraystretch{0.89}
    \setlength{\tabcolsep}{0.9mm}
    \footnotesize
    \caption{\textbf{Monocular 3D object detection on nuScenes}. Fcos3D ranked first among vision-only methods in the NeurIPS 2020 nuScenes 3D Detection Challenge.}
    \vspace{-0.1in}
    
\begin{tabular}{l|c|c}
\hline
    Method & Lr schd &  mAP    \\

    \hline
    \rowcolor{gray!20}
    \textbf{FCOS3D-SW-small}       &   1x   &   31.42   \\ 
    FCOS3D-UniRepLKNet-S           &   1x   &   31.38   \\ 
    FCOS3D-ResNet101 w/ DCN        &   1x   &   29.8    \\
    
    \hline

\end{tabular}
    \label{tab:det3d}
    \vspace{-0.1in}
\end{table}

\begin{figure}
	\begin{center}
		\includegraphics[width=0.99\linewidth]{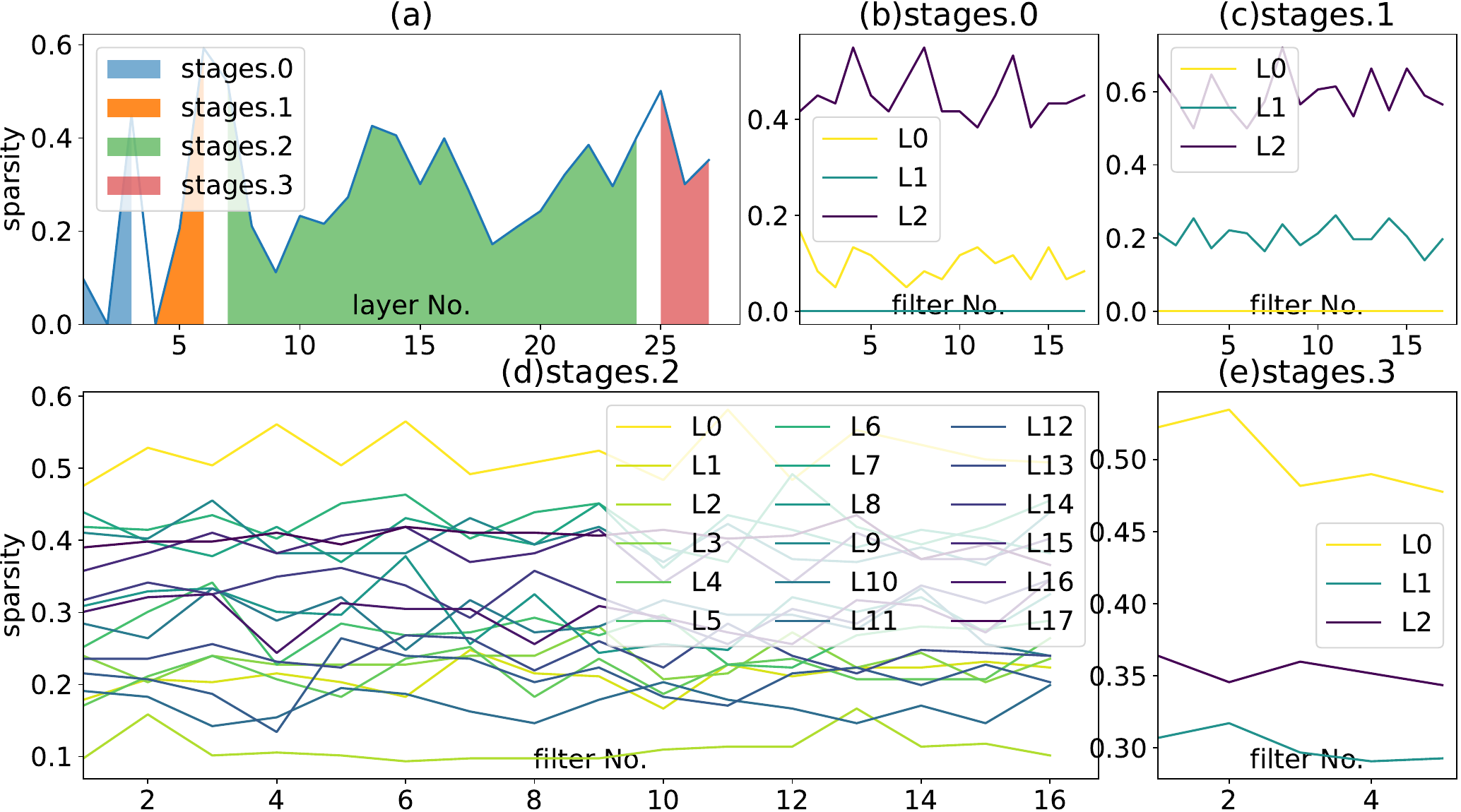}
		\vspace{-2mm}
		\caption{(a) The variation of sparsity with increasing network depth. Colors represent different stages. (b-e) The group convolution of each layer has $\left \lceil \frac{M}{3} \right \rceil$ output channels. The proportion and distribution of the number of removed channel indexes to the total number of groups $nC$. On the four stages, n is [1,2,4,8].}
		\label{fig:sparsity}
		\vspace{-0.35in}
	\end{center}
\end{figure}

\begin{figure}
	\begin{center}
		\includegraphics[width=0.99\linewidth]{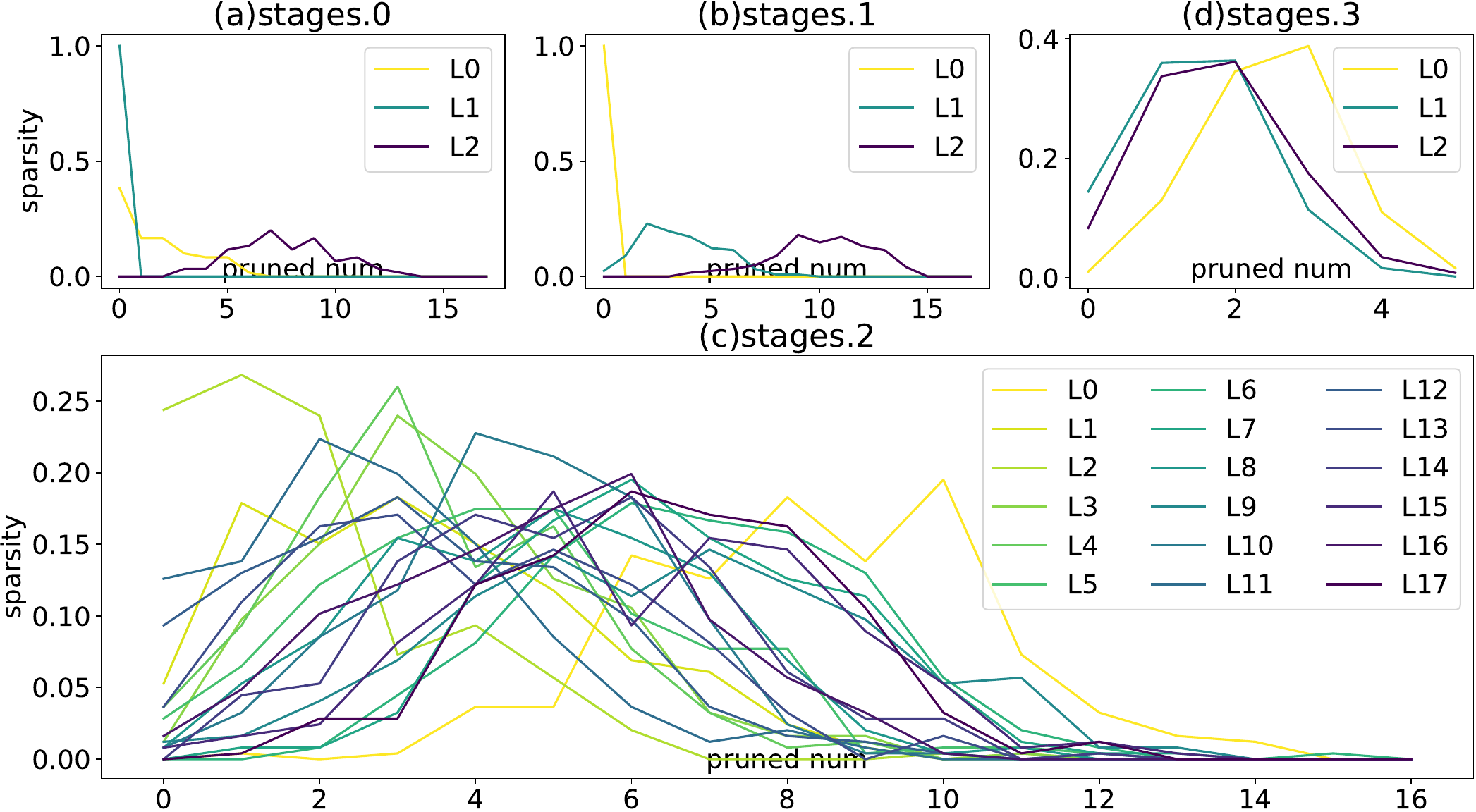}
		\vspace{-2mm}
		\caption{(a-c) There are $nC$ group convolutions in each layer, and the proportion and distribution of groups with the number of filters removed.}
		\label{fig:channelsparsity}
		\vspace{-0.35in}
	\end{center}
\end{figure}

\section{Analysis}
\label{sec:ANALYSIS}
PeLK~\cite{10656330} customizes convolution kernels to mimic the human eye's focus and blur mechanism. This is achieved by managing the parameter count in the central and peripheral regions of the kernel. In contrast to PeLK's manual crafting of large kernel convolutions, our method discovers the spatial structure of convolutions data-drivenly. Our approach uses coarse-grained pruning to eliminate some output channels during training. Unlike traditional network structures, we employ a one-to-many group convolution, with each group having M/3 output channels. For the four stages, the values of M are [51, 49, 47, 13], corresponding to the output channels per group convolution being [17, 17, 16, 5]. Compared to standard network structures, we can remove some output channels from group convolutions before summing them, ensuring the output is minimally affected by pruning. We will analyze the trained model to investigate the data-driven spatial structure of convolutions, focusing on three key aspects: 1. The sparsity variation across network depth. 2. The distribution of pruned convolution locations. 3. The statistical patterns of pruned numbers within a layer.

We define 'sparsity' as the ratio of pruned filters to the total number of filters in each layer and label the network depth index as 'Layer No.' As shown in Fig. \ref{fig:sparsity}~(a), there is a general trend of increased pruning in deeper network layers. Notably, a closer examination at the stage level reveals a significant increase in the number of pruned filters at the last layer of each stage. This rapid increase suggests a reduced demand for complex transformations and a tendency towards information transmission during stage transitions.
Such observations could potentially guide the design of future network architectures. Additionally, UniRepNet employs a strategy that alternates between large and small kernel convolutions across layers. Fig. \ref{fig:sparsity}~(a) also indicates that sparsity varies along the layer index, reflecting UniRep's interlacing strategy, albeit at a lower frequency. Interestingly, reducing the frequency of alternating large kernel usage does not yield significant benefits, a finding that requires further investigation.

We investigated whether the data-driven pruning of large kernel convolutions reveals any clear structural patterns. Our approach started with the collection of indices for all pruned filters across the output channels of each group convolutions, which we labeled as 'Filter No.'. Subsequently, we determined the ratio of pruned filters at each index to the total number of groups involved. Since each group convolution has a single input and multiple outputs, the number of filters at each index before training corresponds to the number of groups. Upon analyzing four stages, we did not identify any distinct patterns. However, a subtle observation was that large and small kernels seem to alternate along the direction of the output channels in group convolutions. This alternation could be a random occurrence or it might indicate the presence of beneficial patterns. Further analysis is required to determine the significance of this observation.

We analyzed the number of pruned filters within each group, which we refer to as 'pruned num'. In a random scenario, the proportion of pruned channels to the total number of group convolutions is [1/17, 1/17, 1/16, 1/5]. Fig. \ref{fig:channelsparsity} illustrates the sparsity after training. Overall, the number of pruned filters aligns with a Gaussian distribution, indicating that a significant number of group convolutions prune a similar quantity of filters at each stage. When comparing across different stages, stages 0-2 exhibit a consistent trend: the number of pruned filters increases with depth. In contrast, stage 3 displays an inverse pattern—where the number of pruned filters decreases as depth increases. Moreover, the initial layers of stages 0 and 1 are more inclined to preserve a greater number of filters. These observations could potentially inform the optimization of hyperparameter settings.

\section{Conclusions}
The progress in CNN models has been significantly influenced by the scaling up of kernel sizes. Our research shifts the focus from kernel size to a decoupled approach, which divides global attention into two critical components: basic granularity feature extraction and feature multi-path fusion. We have demonstrated that standard $3 \times 3$ convolutions can not only match the effects of large convolutions but also further enhance the performance of CNNs. This finding is in resonance with VGG's conclusions. Additionally, our research introduces new parameter settings that were previously unexplored. The analysis of the trained model's parameters has yielded intriguing findings, which could potentially spark further studies in various applications. e.g., Crack Detection \cite{chen2022geometry,chen2023devil,li2025synergistic} and Face recognition \cite{mi2023privacy,mi2024privacy}.
{
    \small
    \clearpage
    \bibliographystyle{ieeenat_fullname}
    \bibliography{main}
}

\clearpage
\setcounter{page}{1}
\maketitlesupplementary

\section*{Appendix A: Inference Throughput Measurement}
\label{sec:throughout} 
In Table \ref{tab:throughput}, we present a comparison of inference throughput. These measurements were conducted on an A100-80GB GPU, using an input resolution of 224 $\times$ 224 pixels. The experiments were carried out with PyTorch version 2.0.0 in conjunction with cuDNN version 11.8.0, employing FP32 precision. As shown in Table \ref{tab:throughput}, our algorithm is notably slower than SLaK~\cite{liu2022more} and UniRepLKNet~\cite{ding2023unireplknet}, with throughput on the A100 GPU at approximately 3/5 and 1/3 of the latter two, respectively.

Utilizing PyTorch's operators, which have not been specifically optimized for large kernel convolutions, our approach yields throughput comparable to SLaK. However, in this scenario, reparameterization (Rep) impedes SLaK's inference speed, reducing throughput by half. This decline is attributed to the merging of strip-like convolutions into a larger convolution, resulting in a substantial number of zero values.

\section*{Appendix B: Operator Optimization}
\label{sec:algorithm} 
Our SW operator possesses four key attributes that facilitate optimization. These characteristics are conducive to enhancing the operator's efficiency.

\subsection*{B.1 Computational Density}
RepLKNet~\cite{ding2022scaling} defends the efficiency of large kernels by demonstrating that small depth-wise (DW) convolutions have a low computation \textit{vs.} memory access cost ratio, which can render DW operations inefficient on GPUs. As kernel size increases, computational density enhances. Our SW operator, as detailed in section ~\cref{sec:architecture}, employs group convolutions that take a single input and yield multiple outputs, matching the computational density of SLaK's strip-like convolutions. Large kernel convolutions, influenced by padding, display reduced computational density at the edges of feature maps, correlating with an arithmetic sequence related to kernel size. In contrast, the small convolutions in our algorithm are less affected by padding's impact on computational density.

\subsection*{B.2 Operator Optimization}
In general, our SW operator capitalizes on the outputs from shared convolutions and stacks them spatially. This spatial stacking operation processes inputs and outputs from the same row or column, which minimizes data loading costs and thereby confers optimization benefits. Specifically, with $E$ denoting the number of edges (section ~\cref{sec:architecture}), each pixel in the input or output undergoes at most $ 2E + 1$ moves. Using atomicAdd for these moves on output pixels can cause delays due to synchronization. However, moving these operations to shared memory can effectively avoid such delays. Moreover, with $E$ defaulted to 4 in this study, the shared memory size required is relatively small.

\subsection*{B.3 Sparsity Gain} 
Our SW algorithm applies coarse-grained pruning to eliminate filters, significantly reducing the computational load during convolutions. Section ~\cref{sec:ANALYSIS} offers an analysis of the resulting sparsity levels. Considering B.1, the operator maintains a high computation \textit{vs.} memory access cost ratio even with sparsity, ensuring a high computational ceiling. Importantly, our approach preserves the module's structure, sidestepping the common issues associated with pruning.

\subsection*{B.4 Operator Fusion} 
The convolution and SW operator are executed sequentially, permitting further operator fusion. This can eliminate the need to transfer data to external memory, using high-speed on-chip storage to complete both operations in one step, conserving VRAM and reducing data loading times.

The version of our SW operator in use is based on atomicAdd and remains unoptimized. The development of an efficient operator that effectively utilizes these features requires the expertise of a skilled CUDA programmer.

\begin{table}[t]
	\centering
	\renewcommand\arraystretch{0.89}
	\setlength{\tabcolsep}{0.9mm}
	\footnotesize
	\caption{\textbf{Inference throughput comparison}. The results are reported in FP32 precision. We conducted this experiment using an A100-80GB GPU with PyTorch 2.0.0 + cuDNN 11.8.0. The symbol $\bigstar$ indicates the use of efficient large-kernel convolutions provided by RepLKNet~\cite{ding2022scaling}. The symbol $\circledR$ denotes Rep.}
	\vspace{-0.1in}
	\begin{tabular}{l|c|c}
		\toprule
			Method & Throughput (img/s) &  IN-1K acc.  \\
			
			\rowcolor{gray!20}
			\midrule
			SW-tiny   &          378    & 83.4\\
			SLaK-T $\circledR\bigstar$ ~\cite{liu2022more}      &   638  & 82.5\\
			UniRepLKNet-T~\cite{ding2023unireplknet}   &          1125    & 83.2 \\
			\rowcolor{gray!20}
			SLaK-T~\cite{liu2022more}      &   371   & 82.5\\
			SLaK-T~ $\circledR$ \cite{liu2022more}      &   174  & 82.5 \\
			
			\rowcolor{gray!20}
			\midrule
			SW-small   &          243    & 83.9 \\
			SLaK-S $\circledR\bigstar$ ~\cite{liu2022more}      &   385   & 83.9\\
			UniRepLKNet-S~\cite{ding2023unireplknet}   &          689     & 83.8 \\
			\rowcolor{gray!20}
			SLaK-S~\cite{liu2022more}      &   217  & 83.8 \\
			SLaK-S~ $\circledR$ \cite{liu2022more}      &   108  & 83.8 \\
			
		\bottomrule
	\end{tabular}
	\label{tab:throughput}
	\vspace{-0.15in}
\end{table}

\section*{Appendix C: Training Configurations}
\label{sec:hyper}
\subsection*{C.1 Phases of Config Evolution}
In experiments \#0 to \#23, we employed the hyperparameters from SLaK, as detailed in Table \ref{tab:appC} under SLaK-T. The top-performing model configuration \#20 achieved a final accuracy of 82.70\% after training for 300 epochs. Upon observing the model's performance at 120 epochs, we suspected overfitting in the 300-epoch training regime. To address this, we selected parameters aimed at mitigating overfitting. We also observed that UniRepLKNet-T's hyperparameters closely resembled those of SLaK-T, albeit with shorter warmup phases and higher dropout rates. Guided by these insights, we combined SLaK-S's sparsity-related hyperparameters with UniRepLKNet-T's to set the configuration for SW-T, as shown in Table \ref{tab:appC}. This approach allowed our SW-T model to outperform both SLaK-T and UniRepLKNet-T, reaching an accuracy of 83.39\%. Notably, with this configuration, the model's accuracy after 120 epochs of training was 82.27\%, the same as \#20. Leveraging this observation, we conducted an experiment (\#25) removing the SE module, akin to \#19 which had achieved 82.25\%. However, in this instance, the accuracy decrease was more pronounced, dropping to 82.14\% (Table \ref{tab:appC300}).

The incorporation of these new hyperparameters introduced some variability in model performance (\#19 \textit{vs.} \#25). However, these differences did not significantly impact the overall conclusions drawn from our trials. The primary objective of our paper is to achieve the effects of large kernel convolutions using small convolutions. Consequently, in addition to the emerging variables (C.2), we continue to utilize the previous parameter configurations and network structures as informed by our experiments. Although our model's hyperparameters are not yet optimal, we have achieved our initial goal. We hope our research will inspire further studies. Detailed training commands for ImageNet-1K are included in the released code.

\subsection*{C.2 Emerging Variables}
Our approach introduces novel variables that have not been previously considered in related research, such as the synchronization frequency of sparse masks and the number of edges in spatially stacked convolutions. These variables significantly influence model performance. For instance, adjusting the initial sparsity values (\#15) has been shown to lead to substantial improvements compared to \#11. The interplay among these variables requires further investigation. Sparse methods illustrate the need for a comprehensive consideration of multiple factors. In our study, experiments manipulating the initial sparsity values (\#15) and the sparsity variations throughout the training process (\#17) both significantly influenced the final model performance. However, these implementations employed different methodologies. This underscores the importance of effectively unifying these elements on a theoretical level. After all, the modified pruning method in our study did not utilize data from the training process but relied on the gradient status at certain steps to influence synchronized sparsity. Such a pruning approach may disproportionately affect small weights near a given threshold, potentially leading to insignificant changes and inconsistent predictions. For details on those parameter settings and their implications, please refer to the training commands on GitHub.

\begin{table}[t!]
	\centering
	\renewcommand\arraystretch{0.89}
	\setlength{\tabcolsep}{0.9mm}
	\footnotesize
	\caption{We trained our model on ImageNet-1K using 8 NVIDIA GTX 4090D GPUs with the following hyperparameters. ``only L'' indicates whether sparsity pruning is applied only to large kernels, ``u'' refers to the update frequency of the sparse mask, ``width factor'' denotes the factor of change in the number of channels in the model, ``magnitude'' represents the pruning process where parameters are sorted by their absolute values and the least significant ones are removed, and ``magnitude\_sum'' means the pruning process where filters are sorted by the sum of their absolute values and the least significant filters are eliminated, respectively.}
	\vspace{-0.1in}
	\resizebox{0.98\linewidth}{!}{
\begin{tabular}{@{\ }l|c|c|c}
\toprule
settings & SLaK-T & SW-tiny & SW-small \\
\midrule
input scale & 
224 & 
224 &
224 \\
batch size & 
4096 & 
4096 &
4096 \\
optimizer &
AdamW & 
AdamW &
AdamW \\
LR      & 
4$\times10^{-3}$ & 
4$\times10^{-3}$ &
4$\times10^{-3}$ \\
LR schedule& 
cosine  &
cosine &
cosine \\
weight decay     &
0.05  & 
0.05  &
0.05 \\
warmup epochs & 
20 &
5 &
5 \\
epochs & 
300 &
300 &
300  \\

\midrule
mixup alpha  & 
0.8 & 
0.8 &
0.8 \\
cutmix alpha &
1.0 & 
1.0 &
1.0 \\
erasing prob. &
0.25    &
0.25   & 
0.25 \\
label smoothing $\varepsilon$ & 
0.1 & 
0.1 &
0.1  \\

\midrule
sparse init & 
snip &
snip &
snip  \\
width factor &
\textbf{1.3} &
\textbf{1.0} &
\textbf{1.0} \\
depths &
[3, 3, 9, 3] &
[3, 3, 18, 3] &
[3, 3, 27, 3] \\
dims &
[96, 192, 384, 768] &
[80, 160, 320, 640] &
[96, 192, 384, 768] \\
kernel size &
[51, 49, 47, 13, \textbf{5}] &
[51, 49, 47, 13, \textbf{3}] &
[51, 49, 47, 13, \textbf{3}] \\
u &
2000 &
100 &
100 \\
prune &
magnitude &
magnitude\_sum &
magnitude\_sum \\
only L &
False &
True &
True \\

\bottomrule
\end{tabular}
}
	\vspace{0.05in}
	\label{tab:appC}
\end{table}

\begin{table}[t!]
	\centering
	\renewcommand\arraystretch{0.89}
	\setlength{\tabcolsep}{0.9mm}
	\footnotesize
	\caption{We conducted analysis experiments to evaluate the model's performance before and after the hyperparameter update. In ablation studies, we analyzed the impact of ghost and sparsity on the model's performance.}
	\vspace{-0.1in}
	\begin{tabular}{l|l|cc}
		\toprule
		&\multirow{2}{*}{SW-tiny} & IN-1K acc. &IN-1K acc.\\
		& &(120 epochs) & (300 epochs)   \\
		
		\midrule
		\#19 & \#20 - SE 		& 82.25 & - \\
		\#20 & the best one & 82.27 & 82.70 \\
		\#24 & u100 		& 82.27 & 83.39 \\ 
		\#25 & \#24 - SE      & 82.14 & - \\ 
		\midrule
		\#26 & \#24 - ghost		& - 	& 82.99 \\ 
		\#27 & sparsity(0.3$\rightarrow$0.2) & - 	& 83.46 \\ 
		\bottomrule
	\end{tabular}
	\vspace{-0.15in}
	\label{tab:appC300}
\end{table}

\begin{figure*}
	\begin{center}
		\includegraphics[width=0.99\linewidth]{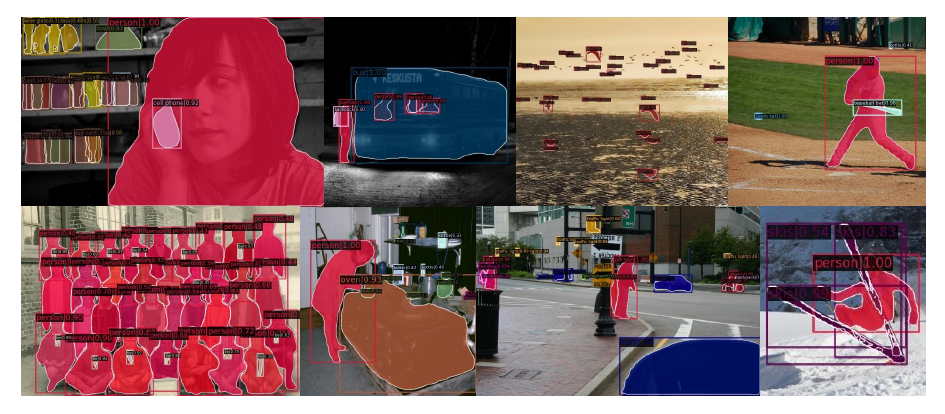}
		\vspace{-0.1in}
		\caption{The qualitative analysis of detection and segmentation results on the COCO dataset. 
        Distinct categories are highlighted with unique colors. This analysis reveals that our method effectively identifies challenging objects, including densely packed entities, those with occlusions, objects in low-light conditions, scenes with significant foreground-background ambiguity, and small objects.}
		\label{fig:vis}
		\vspace{-0.15in}
	\end{center}
\end{figure*}

\begin{figure}
	\begin{center}
		\includegraphics[width=0.99\linewidth]{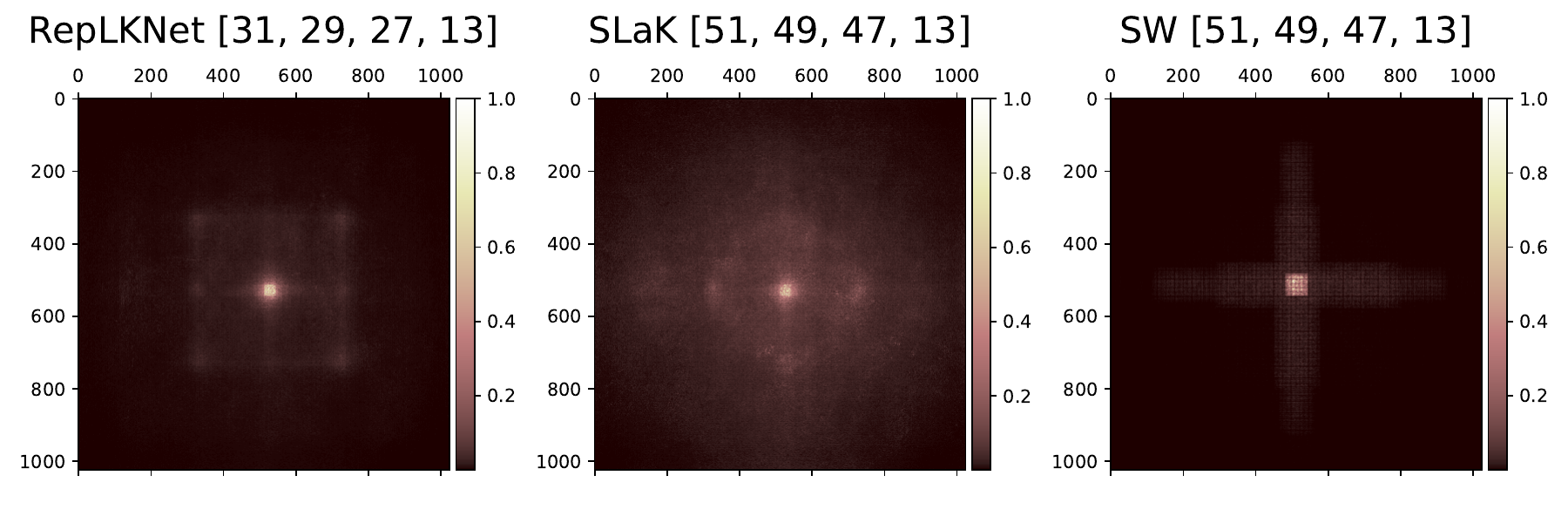}
		\vspace{-2mm}
		\caption{\textbf{Effective receptive fields (ERFs) comparison.}}
		\label{fig:erf}
		\vspace{-0.35in}
	\end{center}
\end{figure}

\section*{Appendix D: Structure Options}
\label{sec:construct}

\noindent
\textbf{Interleaving Large and Small Convolutional Kernels} Experiment \#23, which interleaved large and small convolutional kernels, yielded performance comparable to our final model structure \#20 and reduced the parameter count (Table \ref{tab:appC}). Despite these advantages, we opted not to adopt UniRepLKNet's such structural configuration for three reasons: (1) We aimed to design SW as a plug-and-play module. (2) The equivalent large kernel convolutions in SW consume fewer parameters, eliminating the need to balance their usage with the model's total parameters. (3) Most importantly, the parameters saved were insufficient to significantly increase the model's depth or width.

\noindent
\textbf{Sparsity and Ghost-like Structures} These are key features of the SW architecture, but their necessity is under scrutiny. Both are closely related to our foundational experiments with SLaK. Experiments \#24 and \#26 confirm that ghost structures enhance model performance (Table \ref{tab:appC}). Sparsity, however, is more complex. Early experiments \#15 and \#16 suggest that its role is similar to adjusting the initial sparsity levels across model layers. Additionally, comparing experiments \#24 and \#27 reveals that reducing sparsity can further improve overall model performance (Table \ref{tab:appC}). Given the inherited settings from SLaK and the complexities within sparsity that require further analysis, we have retained sparsity in our experiments for this study.

While these structures indicate that some parameters have not been fully optimized, they have satisfied our initial exploration into convolutional kernel sizes. The pursuit of hyper-parameters to saturate the accuracy of our proposed structure is beyond the scope and resources of this paper.

\section*{Appendix E: Scaling Issues}
\label{sec:scalinglaw}

Our study refrains from testing models with larger parameter and computational demands for three primary reasons: (1) We have inherited hyperparameters from SLaK~\cite{liu2022more} and UniRepLKNet~\cite{ding2023unireplknet}, yet our model is not saturated. There is potential for performance gains through hyperparameter tuning (\#27 \textit{vs.} \#24), which underscores the urgency for increased computational resources. (2) Custom hyperparameters are required for models of varying sizes, and given our method involves more hyperparameters and unoptimized operators, the costs are prohibitive. (3) Our focus is on the scientific exploration of utilizing small convolutions as substitutes for large ones. We present performance benchmarks for models that possess parameter and FLOPs comparable to those of ResNet50 and ResNet101.

\begin{figure}
	\begin{center}
		\includegraphics[width=0.99\linewidth]{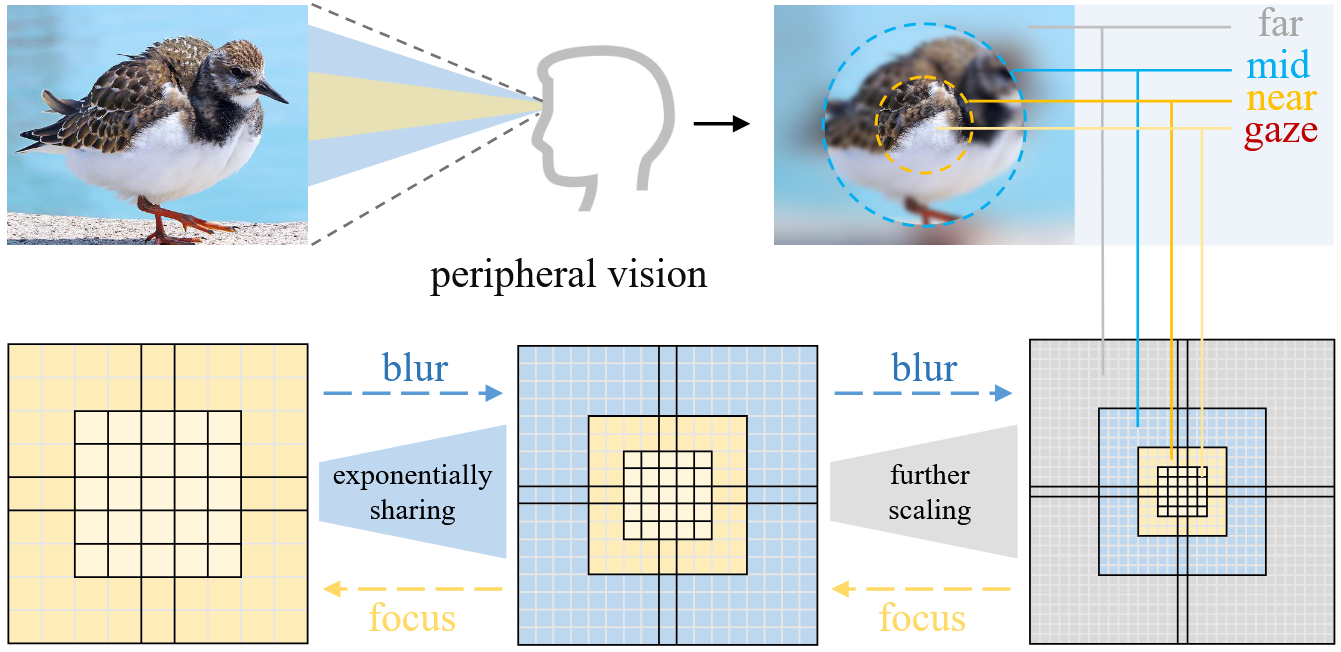}
		\vspace{-2mm}
		\caption{PeLK's~\cite{10656330} focus and blur mechanism, which emulates human peripheral vision. Although our conceptual approach differs from PeLK's, our ERF coincidentally aligns closely with PeLK's mechanisms.
        }
		\label{fig:pelk}
		\vspace{-0.35in}
	\end{center}
\end{figure}

\section*{Appendix F: Analysis}
\label{sec:analysisApp}

\subsection*{F.1 Qualitative Analysis}
We present qualitative analysis results, as shown in Fig. ~\ref{fig:vis}, providing an intuitive understanding of our model's performance on COCO. We selected a subset of challenging images from the COCO dataset and draw the predicted detection boxes and segmentation results on them. It is evident that our model effectively recognizes objects under various lighting conditions, across different object sizes, amidst occlusions, and against complex backgrounds. The segmentation results feature well-defined visual boundaries, like mobile phones and skateboards. The SW operator demonstrates robust marginal detection capabilities in these complex tasks. Such performance lays a solid foundation for the future integration of this module into self-supervised learning or sophisticated generative tasks.

\subsection*{F.2 Visualization of ERFs.}
In accordance with the method used by RepLKNet~\cite{ding2022scaling} and SLaK~\cite{liu2022more} for generating Effective Receptive Fields (ERFs), we have obtained a 1024$\times$1024 matrix, as illustrated in Fig.~\ref{fig:erf}. For further details on the generation process, refer to SLaK~\cite{liu2022more}. Our ERFs contrasts with theirs by displaying robust correlations within the immediate neighborhood that diminish progressively with distance. Our ERFs pattern is more predictable, which better complements the sliding window mechanism of convolutions. Despite our conceptual divergence from PeLK~\cite{10656330} and the resulting disparity in convolutional design, our ERFs closely mirrors PeLK's focus and blur mechanisms (Fig.~\ref{fig:pelk}).

\end{document}